\documentclass[10pt,twocolumn,letterpaper]{article}

\usepackage{iccv}
\usepackage{times}
\usepackage{epsfig}
\usepackage{graphicx}
\usepackage{amsmath}
\usepackage{amssymb}
\usepackage{subcaption}
\usepackage[skip=5pt]{caption}
\usepackage{multirow}
\usepackage{booktabs}
\usepackage{siunitx}

\usepackage{pdfpages}

\usepackage{url}

\usepackage[pagebackref=true,breaklinks=true,letterpaper=true,colorlinks,bookmarks=false]{hyperref}

 \iccvfinalcopy % *** Uncomment this line for the final submission

\def\iccvPaperID{****} % *** Enter the ICCV Paper ID here

\ificcvfinal\fi

\def\numVideo{22,075}
\def\numFrame{3,735,476}   % To be revised before public release
\def\numSub{3,107}
\def\numBatchOne{140,000}

\title{The Do's and Don'ts for CNN-based Face Verification}

\author{Ankan Bansal
	\qquad
	Carlos Castillo
	\qquad
	Rajeev Ranjan
	\qquad
	Rama Chellappa\\
	UMIACS\\
	University of Maryland, College Park\\
	\texttt{\{ankan,carlos,rranjan1,rama\}@umiacs.umd.edu}
}

\begin{document}
	
	\maketitle
	%\thispagestyle{empty}

	%%%%%%%%% ABSTRACT
	\begin{abstract}
		While the research community appears to have developed a consensus on the methods of acquiring annotated data, design and training of CNNs, many questions still remain to be answered. In this paper, we explore the following questions that are critical to face recognition research: (i) Can we train on still images and expect the systems to work on videos? (ii) Are deeper datasets better than wider datasets? (iii) Does adding label noise lead to improvement in performance of deep networks? (iv) Is alignment needed for face recognition? We address these questions by training CNNs using CASIA-WebFace,  UMDFaces, and a new video dataset and testing on YouTubeFaces, IJB-A and a disjoint portion of UMDFaces datasets. Our new data set, which will be made publicly available, has \numVideo~videos and \numFrame~human annotated frames extracted from them.
	\end{abstract}
	
	%%%%%%%%% BODY TEXT
	\section{Introduction}
	The re-emergence of deep convolutional neural networks has led to large improvements in performance on several face recognition and verification datasets \cite{lfw, lfwsurvey, ijba}. However, the face recognition problem isn't ``solved" yet. The process of training a face recognition system starts with choosing a dataset of face images, detecting faces in images, cropping and aligning these faces, and then training deep networks on the cropped and possibly aligned faces. Every step of the process involves many design issues and choices. 
	
	Some issues have received significant attention from researchers. These include choices about the architecture of neural networks. On the other hand, there are several other design choices which require more attention. These arise at every stage of the process from face detection and thumbnail (image obtained after cropping and aligning the face image) generation to selecting the training dataset itself. We tackle some of these design questions in this paper. We also introduce a dataset of \numVideo~videos collected from YouTube of \numSub~subjects. These subjects are mainly from batch-1 of the recently released UMDFaces \cite{umdfaces} dataset. We release face annotations for \numFrame~frames from these videos and the corresponding frames separately. We use this dataset to study the effect of using a mixture of video frames and still images on verification performance for unconstrained faces such as the IJB-A \cite{ijba} and YTF \cite{ytf} sets. 
	
	Face detection is the first step in any face recognition pipeline. Several CNN-based face and object detectors have been introduced which achieve good detection performance and speeds \cite{rajeev-btas, faster-rcnn, yolo, ssd, hyperface, cms-rcnn, ultraface, tiny}. Each of these detectors learns a different representation. This leads to generation of different types of bounding boxes for faces. Verification accuracy can be affected by the type of bounding box used. In addition, most recent face recognition and verification methods \cite{deepface, swami, deepid_3, jc15, frankenstein, hybrid} use some kind of 2D or 3D alignment procedure \cite{unconstrained, dlib, ultraface, frontalization}. All these variables can lead to changes in performance of deep networks. To the best of our knowledge there has been very little systematic study of effects of the thumbnail generation process \cite{vgg} on the accuracy of deep networks. In section \ref{sec:myth4} we study the consequences of using different thumbnail generation methods. We show that using a good keypoint detection method and aligning faces both during training and testing leads to the best performance. 
	
	Other questions concern the dataset collection and cleaning process itself. The size of available face datasets can range from a few hundred thousand images \cite{umdfaces,casia,facescrub, ytf} to a few million \cite{megaface1,megaface2,msceleb,msceleb1,vgg}. Other datasets, which are not publicly available, can go from several million faces \cite{deepface} to several hundred million faces \cite{facenet}. Much of the work in face recognition research might behave differently with such large datasets. Apart from datasets geared towards training deep networks, some datasets focus on evaluation of the trained models \cite{lfw, ijba}. All of these datasets were collected using different methodologies and techniques. For example, \cite{ytf} contains videos collected from the internet which look quite different from still image datasets like \cite{casia, umdfaces, msceleb}. We study the effects of this difference between still images and frames extracted from videos in section \ref{sec:myth1} using our new dataset. We found that mixing both still images and the large number of video frames during training performs better than using just still images or video frames for testing on any of the test datasets \cite{ijba,ytf,umdfaces}. 
	
	In section \ref{sec:myth2}, we investigate the impact of using a deep dataset against using a wider dataset. For two datasets with the same number of images, we call one deeper than the other if on average it has more images per subject than the other. We show that the choice of the dataset depends on the kind of network being trained. Deeper networks perform well with deeper datasets and shallower networks work well with wider datasets.
	
	Label noise is the phenomenon of assigning an incorrect label to some images. Label noise is an inherent part of the data collection process. Some authors intentionally leave in some label noise \cite{vgg, msceleb1, msceleb} in the dataset in hopes of making the deep networks more robust. In section \ref{sec:myth3} we examine the effect of this label noise on the performance of deep networks for verification trained on these datasets and demonstrate that clean datasets almost always lead to significantly better performance than noisy datasets.
	
	We make the following main contributions in this paper:
	\begin{itemize}
		\setlength\itemsep{-0.4em}
		\item We conduct a large scale systematic study about the effects of making certain apparently routine decisions about the training procedure. Our experiments show that data diversity, number of individuals in the dataset, quality of the dataset, and good alignment are keys to obtaining good performance.
		%	\item We show that careful consideration before taking some apparently mundane decisions can lead to significant jumps in verification performance.
		\item We suggest some general rules that could lead to improvement in the performance of deep face recognition networks. These practices will also guide future data collection efforts.
		\item We introduce a large dataset of videos of over 3,000 subjects along with \numFrame~human annotated bounding boxes in frames extracted from these videos.
	\end{itemize}
	
	%%%%%%%%%%%%%%%%%%%%%%%%%%%%%%%%%%%%%%%%%%%%%%%%%%%%%%%%%%%%%%%%%%%
	\section{Dataset}
	\label{sec:dataset}
	%Hundreds of hours of videos are uploaded to YouTube every minute. 
	Still photos from the internet cannot match the amount of variation that videos provide. Videos (and frames extracted from the videos) are under-utilized because of the difficulty in cleaning and annotating the data. There is a need for effective methods for annotating video data. 
	We describe a new dataset\footnote{\url{http://umdfaces.io/}} aimed at face recognition research. It contains \numVideo~videos of \numSub~subjects collected from YouTube. We provide bounding box annotations for \numFrame~frames from the videos. We explain our methodology of collecting this dataset which, we hope, will be useful to researchers working on face verification and related problems.
	
	\begin{figure*}
		\begin{center}
			\includegraphics[width=0.7\linewidth,height=0.3\linewidth]{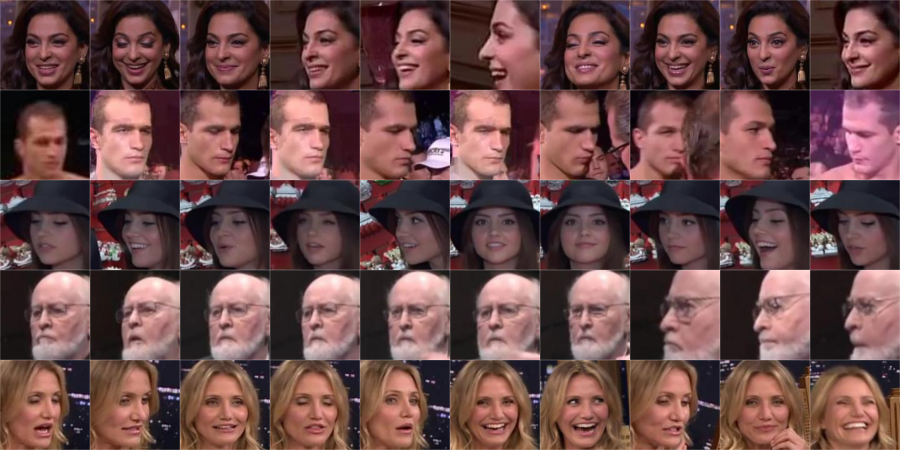}
			%		\fbox{\rule{0pt}{2in} \rule{.9\linewidth}{0pt}}
		\end{center}
		\caption{Some sample annotated bounding boxes from the dataset. Each row contains frames from a video. There is a large amount of pose and expression variation in each video.}
		\label{fig:dataset}
	\end{figure*}
	
	\subsection{Collecting data}
	We searched YouTube for over 3000 subject identities (from batch-1 of UMDFaces \cite{umdfaces}) and tried to download the first 20 videos for each person. We used the open source system youtube-dl \cite{youtube-dl} for searching and downloading the videos. We downloaded a total of about 40,000 videos. 
	
	\subsection{Automated filtering}
	From each video, we extracted either all the frames or the first 4,000 frames, whichever is lower. This process gave us over 140 million frames. We randomly selected about 10\% of these frames to process further. Next we detected faces in the retained frames. At the time of collecting this data, most detection systems were too slow to be of use. Faster RCNN \cite{faster-rcnn} claimed to detect objects at 7 frames per second. We decided to use the, then newly introduced, YOLO detector \cite{yolo}. We trained the YOLO detector on the WIDER dataset \cite{wider} and fine-tuned on the FDDB dataset \cite{fddb}. We were able to run the trained detector at about 25 frames per second on a Titan X GPU. This enabled us to detect all faces in these 14 million images in less that one week. This gave us over 40 million face boxes in 14 million frames. We again randomly selected 4000 face boxes for each subject identity finally leaving ourselves with about 14 million boxes.
	
	Our next task was to remove all face box proposals which did not belong to the person in question. We used the all-in-one method proposed in \cite{ultraface} to detect key landmark points on each face and used them to align the faces. We used the images in batch-1 of the UMDFaces dataset \cite{umdfaces} as reference images for the subjects. Our problem now reduced to a verification problem. For each subject we need to verify whether a face box belongs to that person. 
	
	We used the verification method proposed in \cite{jc15} for filtering the proposal boxes which are not of the person in question. We extracted features (using a network trained in the same way as \cite{casia,jc15}) for all images in batch-1 of UMDFaces \cite{umdfaces} and take their average over a subject to obtain one feature vector for the subject. Then, for each face box in our dataset for the subject, we compared the feature vector with the reference feature vector obtained above and kept only those boxes with similarity with the reference feature vector above a threshold. We used cosine similarity as the similarity metric and used a low threshold to avoid removing the hard-positive examples from the dataset as these are very valuable. This leaves us with about 4 million face boxes.% To further clean these and be more confident about the identity of the subjects in these boxes we crowd-source the task of filtering the proposals.
	
	\subsection{Crowd-sourcing final filtering}
	To obtain the final dataset, we use Amazon Mechanical Turk (AMT) to filter the proposals. We show each proposal to 2 `mechanical turkers'. Each screen in our AMT task contains 50 images to be filtered and 3 reference images of a subject obtained from the UMDFaces dataset \cite{umdfaces}. We requested the mechanical turkers to select images which do not belong to the subject under consideration. We removed all the faces boxes which were selected by at least one turker. To ensure high quality annotations, we adopted the following quality control method.
	
	\subsection{Quality control through sentinels}
	We used the method similar to the one used in \cite{bala,umdfaces} for controlling the quality of annotations. Each screen of 50 images contains 5 known images of another subject. Depending on whether the turkers select these `sentinel' images, they get an accuracy score. We only considered the votes of turkers with high accuracy scores.
	
	\subsection{Result}
	After the final filtering through human annotators, we have \numFrame~annotated frames in \numVideo~videos. We will publicly release this massive dataset for use by the computer vision community.
	
	%%%%%%%%%%%%%%%%%%%%%%%%%%%%%%%%%%
	
	Next, we use this dataset to show the importance of using video frames for training while testing on real world scenarios like IJB-A and YTF in section \ref{sec:myth1}. We show that utilizing the vast amounts of video data and mixing video frames with still images can give a significant boost in verification performance over using only video frames or still images. 
	
	%-------------------------------------------------------------------------
	\section{Questions and Experiments}
	\label{sec:myths}
	We show that judicious decisions about the training set and procedures can lead to large improvements in verification accuracy of deep networks. We first use the introduced dataset to show the importance of using video frames while training for verification. Then we investigate some more questions that will guide researchers towards good practices for training deep networks for face verification and identification. These include: (i) whether deep datasets are better than wide datasets (section \ref{sec:myth2}); (ii) whether label noise helps in improving performance (section \ref{sec:myth3}); and (iii) how important is the thumbnail generation method for training and testing deep networks (section \ref{sec:myth4}). We use the Caffe \cite{caffe} framework for all experiments.
	
	\subsection{Do deep recognition networks trained on stills perform well on videos?}
	%\subsection{Do deep recognition networks trained on still images perform well on videos and video frames?}
	\label{sec:myth1}
	Images in most still image datasets \cite{lfw, celebA, pubfig} are taken with high quality cameras in good lighting. Photos of celebrities on the internet are often selected from among several taken by a professional photographer.  This introduces a bias towards high quality images. Models trained on only still images perform poorly on frames extracted from videos \cite{ijba}. These frames are extremely challenging because of pose, expression, and lighting variations. At the same time, models trained only on videos perform poorly on still images. There is a huge amount of video data available and only a limited number of still images. We show that training on a mixture of images and video frames is really important for achieving good verification performance.
	
	We train deep networks on the following five sets and compare the verification performance of these networks:
	\begin{itemize}
		\setlength\itemsep{-0.4em}
		\item \textbf{Stills}: Some part (batch-1) of the UMDFaces \cite{umdfaces} dataset. This comprises of about \numBatchOne~still images. We train an Alexnet-derived architecture \cite{swami} on these images for 100,000 iterations with a batch size of 128 and initial learning rate of 0.01 and reduced by half every 15,000 iterations. 
		\item \textbf{Frames}: The same number (\numBatchOne) of video frames from our dataset (section \ref{sec:dataset}). Each subject has the same number of images as the case above (Stills). We used the same training method as above.
		\item \textbf{Frames++}: The same number of subjects as above but using many more video frames per subject for a total of about 1 million video frames. We trained this model for 100,000 iterations and decreased the learning rate by half every 20,000 iterations. 
		\item \textbf{Mixture}: A mixture of still images and video frames from UMDFaces and our dataset. We took 50\% of images from batch-1 of UMDFaces and the other 50\% from our video frames dataset for a total of about \numBatchOne~images. We trained this network for 100,000 iterations.
		\item \textbf{Mixture++}: The same number of still images as `Stills' but about 1 million video frames. We again train the network on this dataset for 100,000 iterations.
	\end{itemize}
	
	Note that we are using far more images in the Frames++ and Mixture++ cases than the other cases. However, we believe that it is fair to compare these five methods because it is much easier to obtain millions of video frames than to obtain millions of still images. There is a lot more variation in 100 images than there is in 100 continuous video frames. Also, in real world scenarios, the amount of video data is increasing rapidly and the majority of recognition has to happen in videos. 
	
	We use an  architecture \cite{swami} derived from Alexnet \cite{alexnet} due to it's easy availability and practicality. It is very fast to train and is perfectly suited for large-scale experiments like ours. Also, it provides excellent results \cite{swami}. However, we believe that, in this case, our observations are general and will be valid for other network architectures too. 
	
	We trained networks on these five sets and compare performance of the trained models on IJB-A \cite{ijba}, YouTube Faces datasets \cite{ytf} and batch-3 of the UMDFaces \cite{umdfaces} dataset. In this experiment and the rest of the paper, unless otherwise stated, we train the same architecture of networks on different datasets for a fixed number of iterations (100,000). We adopt the 1:1 verification protocol similar to the one introduced in \cite{ijba} for evaluating the performance of these deep networks. We give a brief description of the evaluation protocol next. 
	
	\subsubsection{Protocol}
	\label{sec:protocol}
	The IJB-A 1:1 verification protocol \cite{ijba} uses a decision error tradeoff (DET) curve for evaluation. The DET curve is equivalent to an ROC curve. %which plots the true positive match rate against the false positive match rate as a function of similarity threshold for a set of pairs of templates (groupings of images and/or frames for the same subject) for verification \cite{vgg}. 
	In our examples we evaluate the performance for 1:1 verification on pairs of images or templates for different datasets \cite{lfw,ytf,umdfaces}. For all experiments, we use ROC curves for evaluation.   
	
	\subsubsection{Results}
\begin{figure}[t]
		\begin{center}
			\includegraphics[width=\linewidth,height=0.7\linewidth]{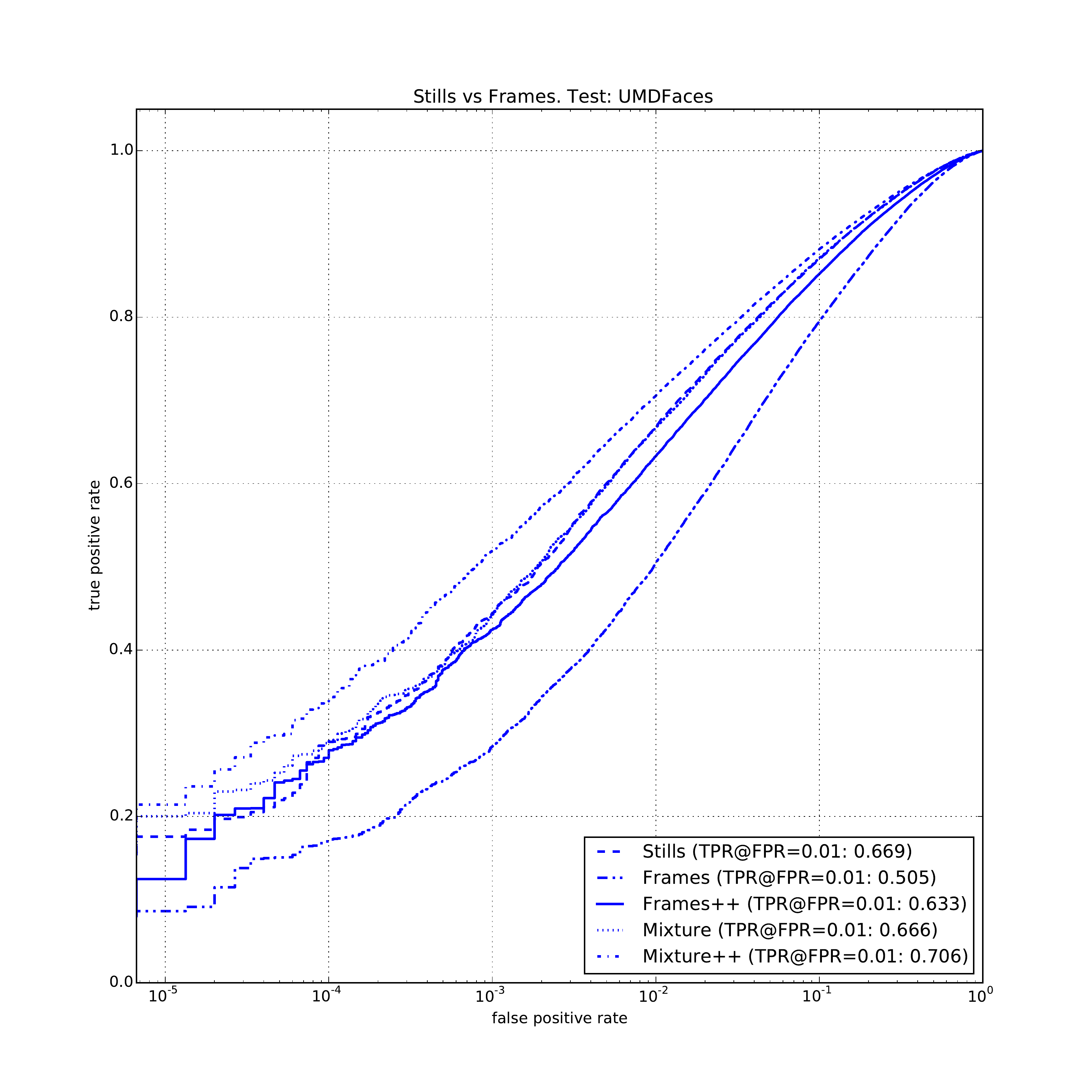}
		\end{center}
		\caption{Verification performance of networks trained on `Stills', `Frames', `Mixture', `Frames++', `Mixture++' and tested on UMDFaces batch-3 \cite{umdfaces}. Note that the test set comprises of only still images. The performance of `Stills' and `Mixture' is very similar. However, `Mixture++' performs best. `Stills' performs the next best after `Mixture++' in this case.}
		\label{fig:myth1_umd}
	\end{figure}

\begin{figure}[t]
		\begin{center}
			\includegraphics[width=\linewidth,height=0.7\linewidth]{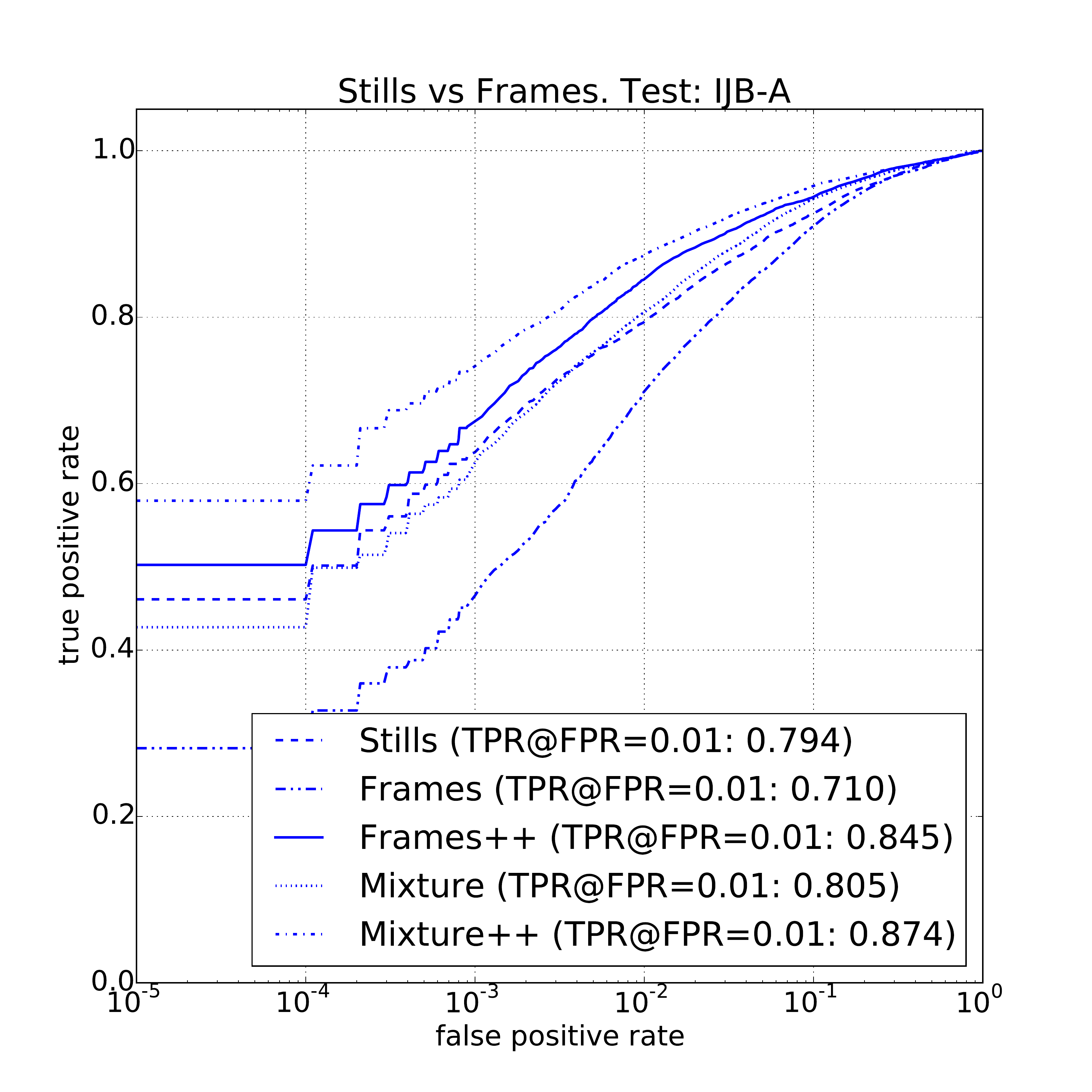}
		\end{center}
		\caption{Verification performance of the five networks (Stills, Mixture, Frames, Mixture++, and Frames++) on IJB-A test set \cite{ijba}. The IJB-A test set contains a mixture of still images and video frames. Again, the performance of `Stills' and `Mixed' are almost the same and `Mixture++' is better than everything else. However, unlike figure \ref{fig:myth1_umd}, the performance of `Frames++' is better than `Stills'.}
		\label{fig:myth1_ijba}
	\end{figure}

	Figures \ref{fig:myth1_umd} and \ref{fig:myth1_ijba} show the performance of the above five experiments. They clearly show the importance of using a mixture of video frames and still images for all cases. We see that while the performance of the `Stills' and `Mixture' cases is very close for both IJB-A and UMDFaces, the performance of `Frames' is very poor. This is because of the presence of many still images in the test sets and the low variety in the few training frames. On the other hand, note that the performance of the `Mixture++' case is much better than any other case, even better than `Frames++' which has similar number of images. This shows the importance of using both still images and the ample number of frames extracted from videos for improving verification performance on unconstrained faces.
	
\begin{figure}[t]
		\begin{center}
			\includegraphics[width=\linewidth,height=0.7\linewidth]{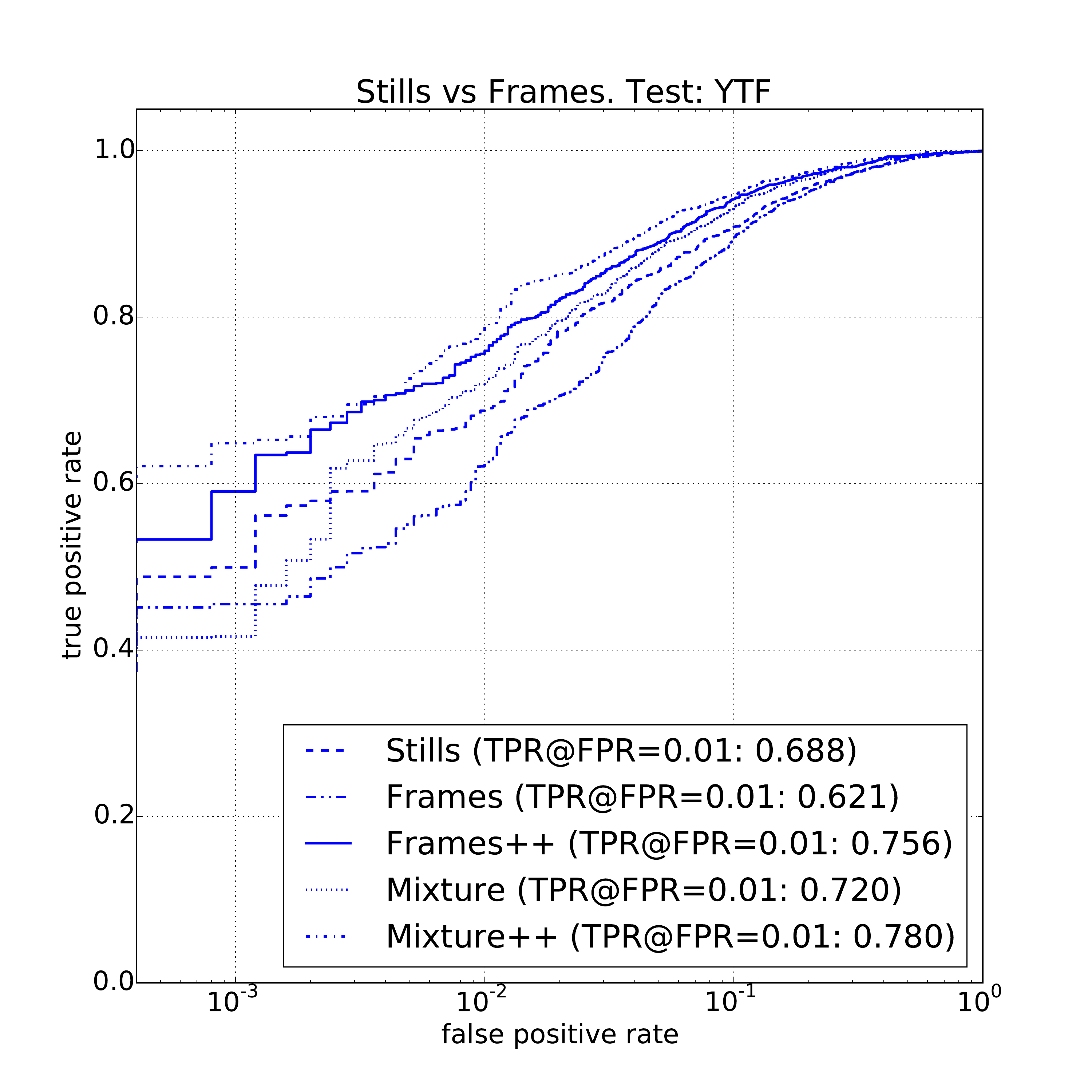}
		\end{center}
		\caption{Verification performance of the five networks (Stills, Mixture, Frames, Mixture++, and Frames++) on YTF test set \cite{ytf}. The test set contains only frames extracted from videos. Again, the performance of `Mixture++' is better than everything else. Also, `Mixture' performs better than `Stills' in this case.}
		\label{fig:myth1_ytf}
	\end{figure}

	Also, note from figure \ref{fig:myth1_ijba} that when testing on a dataset which contains a mixture of still images and video frames \cite{ijba}, the performance of `Mixture++' is the highest and `Frames++' is the second highest. However, when testing on the UMDFaces dataset \cite{umdfaces} which contains only images, `Stills' performs second best after `Mixture++' (figure \ref{fig:myth1_umd}). Similarly, when testing on the completely video-based testing set YTF \cite{ytf}, from figure \ref{fig:myth1_ytf}, `Mixture++' performs the best and `Frames++' performs a bit worse than it. Also note that `Mixture' performs better than `Stills' and `Frames'. Collecting millions of still images with enough variations is extremely difficult. It is much easier to collect and annotate millions of video frames. Also, using a combinations of a large number of video frames and relatively few still images gives a significant boost in performance over using only still images or video frames.
	
	\subsection{What is better: deeper or wider datasets?}
	\label{sec:myth2}
	For datasets with the same number of total (still) images, we call a dataset with more images per subject deeper than another dataset with fewer images per subject. We call the latter dataset wider than the prior. An example of a deep (deeper than many other still image datasets) dataset is the VGG-Face dataset \cite{vgg} which has about 2.6 million images of 2,622 subjects. On the other hand CASIA-WebFace\cite{casia} can be considered a wide dataset. An extreme example of a wide dataset is the MegaFace training dataset \cite{megaface1, megaface2} which has over 670,000 subjects and only about 7 images per subject.
	
	It is not intuitively clear whether it's better to use deeper datasets or wider for training deep networks. Given enough images, both deep and wide datasets can contain a variety of face images. Deep datasets are more varied in pose, expression, illuminations etc. On the other hand wide datasets contain large variations because of the large number of unique identities. In this section, we try to resolve the dilemma of choosing one kind of dataset over the other.
	
	We use the UMDFaces \cite{umdfaces}, MS-Celeb-1M \cite{msceleb} and CASIA-WebFace \cite{casia} datasets to analyze the question. We treat batch-1 and batch-2 of UMDFaces as the training set. To explore the question of deeper vs wider datasets, we divide the training datasets into two as follows: We sort the subjects according to the number of images they have; then we start with the subject with the maximum number of images and put the subject in  one set (head); we then take the subject with next highest number of images and add him/her to the head set; we continue this process till we have collected close to half the total number of images. Now we have divided each dataset into two parts. The first part (which we call `head') contains the deeper half of the dataset. The other half is called the `tail'. For CASIA-WebFace, the `head' dataset contains 1,738 subjects and 247,196 images and the `tail' set contains 8,437 subjects and 247,218 images. Similarly, the UMDFaces `head' set has 2,142 subjects with 144,371 images and the `tail' set has 4,092 subjects and 144,348 images. 
	
\begin{figure}[t]
		\begin{center}
			\includegraphics[width=\linewidth,height=0.7\linewidth]{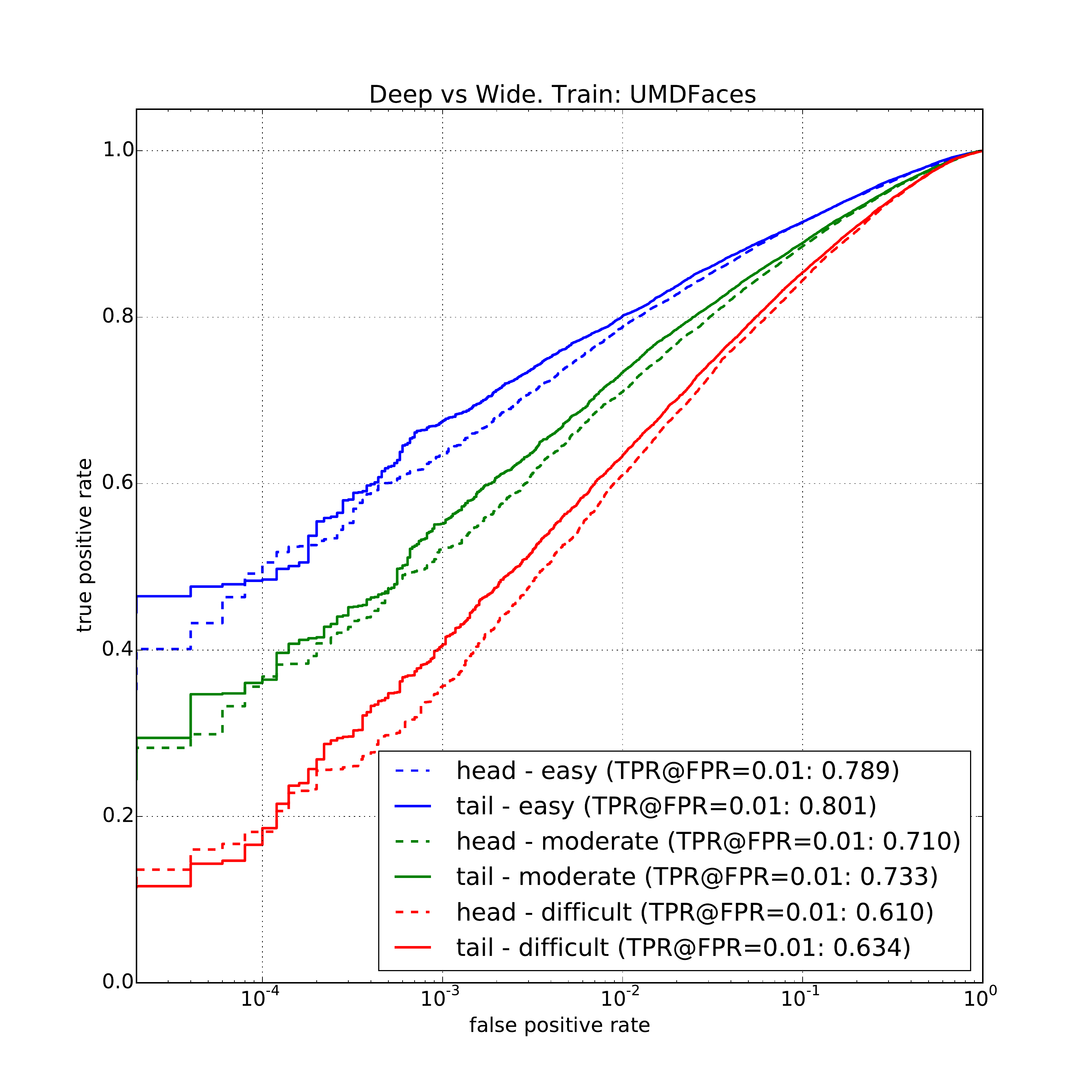}
		\end{center}
		\caption{Training on UMDFaces \cite{umdfaces} batch-1 and batch-2 and testing on batch-3. Solid lines represent training on the `tail' (wide) set and dashed lines represent training on the `head' set. We show the performance over three parts of the test dataset: easy, moderate, and hard. These parts are based on the difference in pose of the pair of images. The performance of the network trained on the `tail' set is invariably better.}
		\label{fig:myth2_umd_umd}
	\end{figure}

\begin{figure}[t]
		\begin{center}
			\includegraphics[width=\linewidth,height=0.7\linewidth]{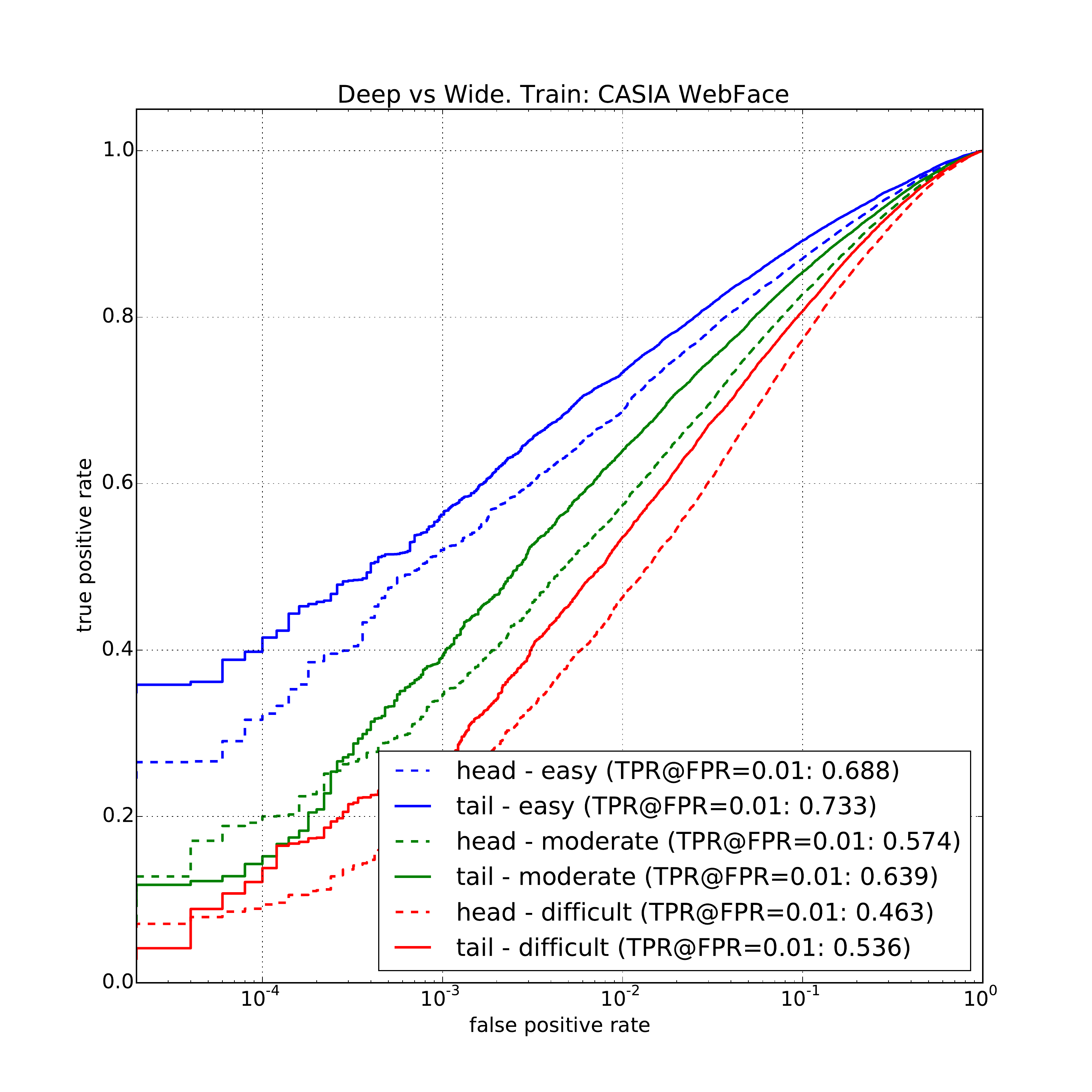}
		\end{center}
		\caption{Verification performance of the networks trained on CASIA-WebFace \cite{casia} `head' and `tail' sets. We see similar trends as figure \ref{fig:myth2_umd_umd}.}
		\label{fig:myth2_casia_umd}
	\end{figure}

\begin{figure}[htp]
		\centering
		\begin{subfigure}{.5\linewidth}
			\centering
			\includegraphics[width=\linewidth,height=0.8\linewidth]{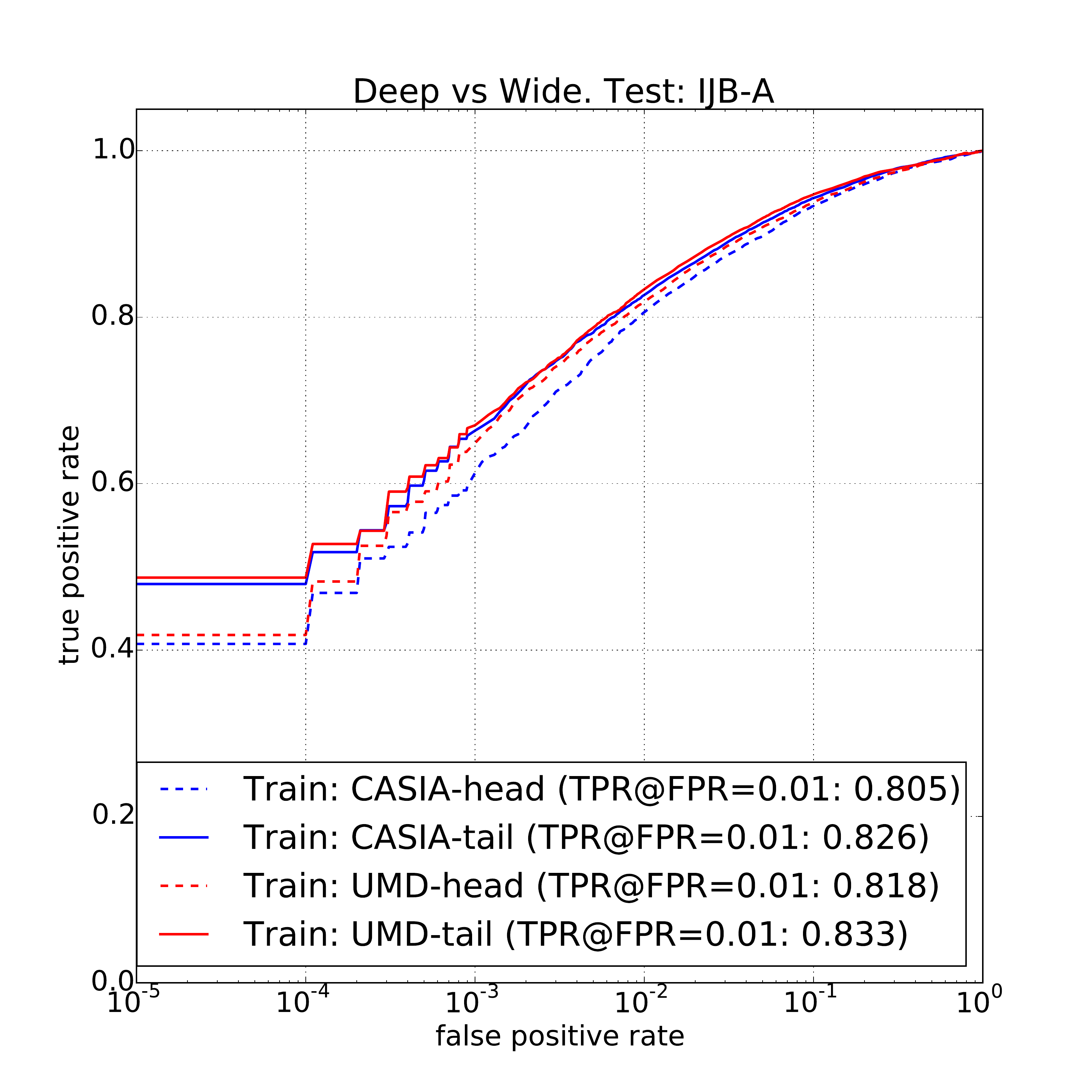}
			\caption{IJB-A}
			\label{fig:myth2_ijba}
		\end{subfigure}%
		\begin{subfigure}{.5\linewidth}
			\centering
			\includegraphics[width=\linewidth,height=0.8\linewidth]{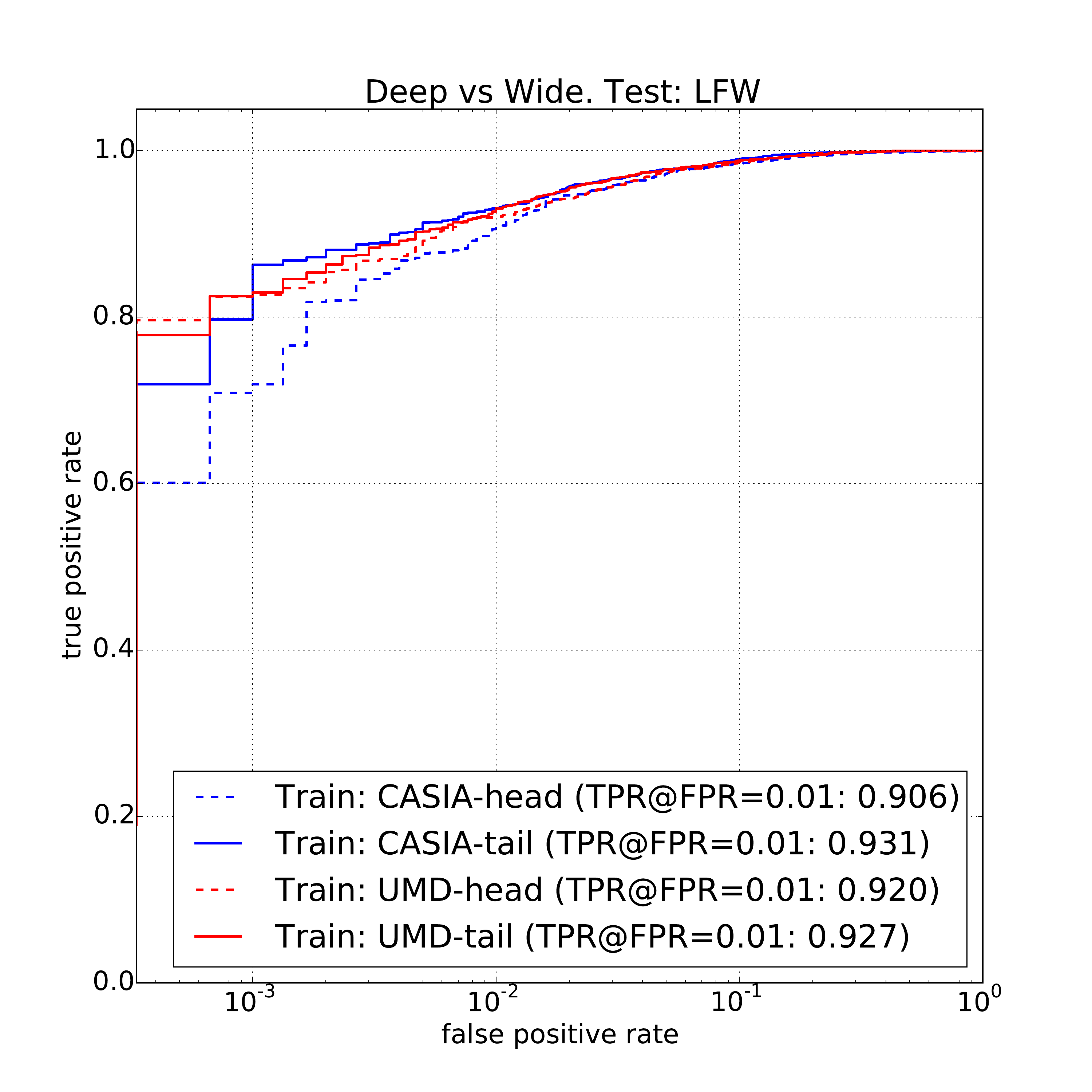}
			\caption{LFW}
			\label{fig:myth2_lfw}
		\end{subfigure}
		\caption{Performance on (a) IJB-A \cite{ijba} and (b) LFW \cite{lfw} of the networks trained on CASIA \cite{casia}, and UMDFaces \cite{umdfaces} `head' and `trail' sets. The performance of the networks trained on `tail' are better across the range of false positive rate. (Best viewed digitally)}
		\label{fig:myth2_ijba_lfw}
	\end{figure}

	We first train the same architecture networks on the `head' and `tail' sets of both CASIA-WebFace and UMDFaces. We test these networks using the protocol from section \ref{sec:protocol} on the UMDFaces batch-3 \cite{umdfaces}, IJB-A \cite{ijba}, and Labeled Faces in the Wild (LFW) \cite{lfw} datasets. The results are shown in figures \ref{fig:myth2_umd_umd}, \ref{fig:myth2_casia_umd}, and \ref{fig:myth2_ijba_lfw}. We note that the performance of the network trained on the `tail' sets is better than the corresponding network trained on the `head' sets for all three test sets. This means that, for a given number of images, it is better to have more subjects than having more images for fewer subjects.
	
\begin{figure}[htp]
		\centering
		\begin{subfigure}{.5\linewidth}
			\centering
			\includegraphics[width=\linewidth,height=0.8\linewidth]{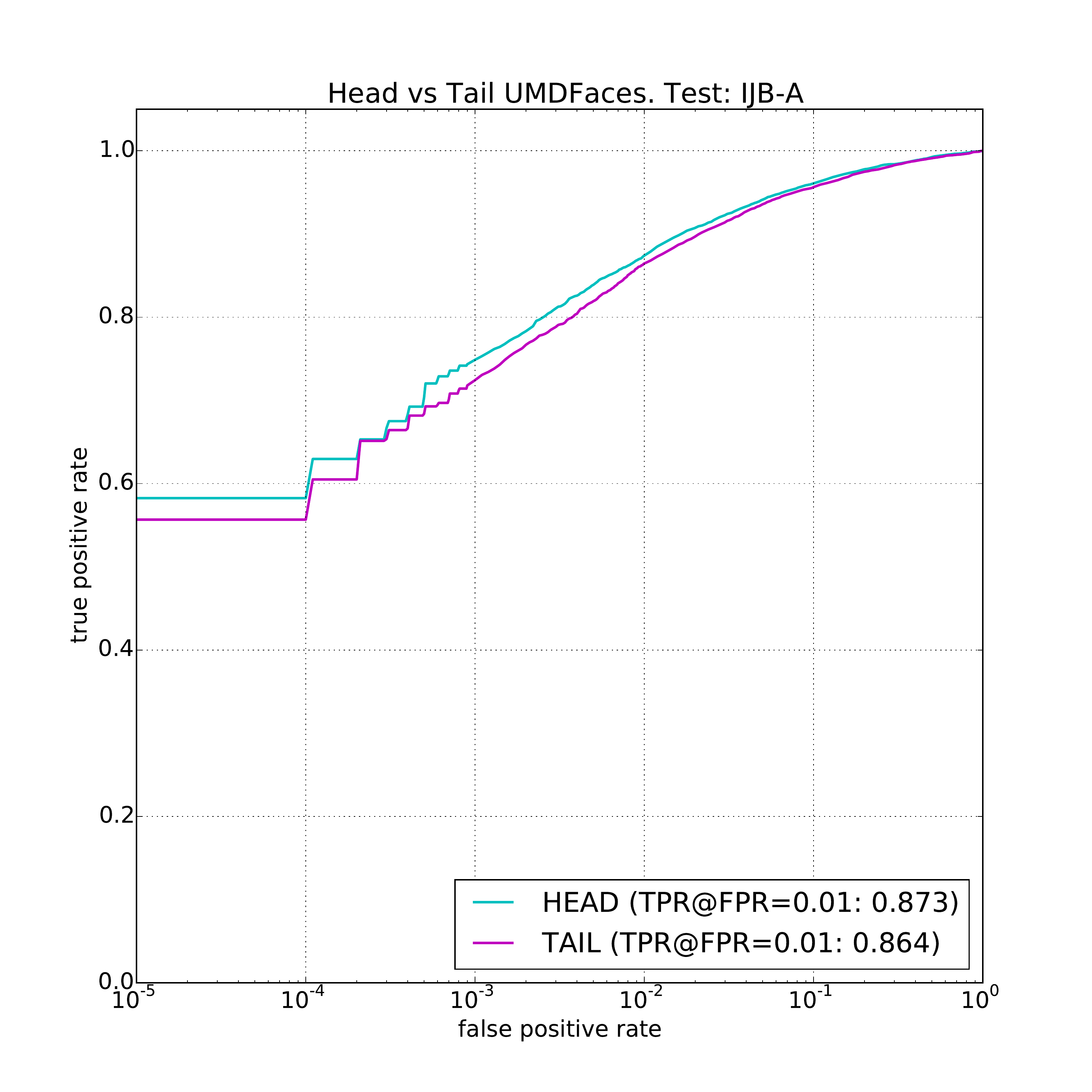}
			\caption{UMDFaces}
			\label{fig:myth2_resnet_umd}
		\end{subfigure}%
		\begin{subfigure}{.5\linewidth}
			\centering
			\includegraphics[width=\linewidth,height=0.8\linewidth]{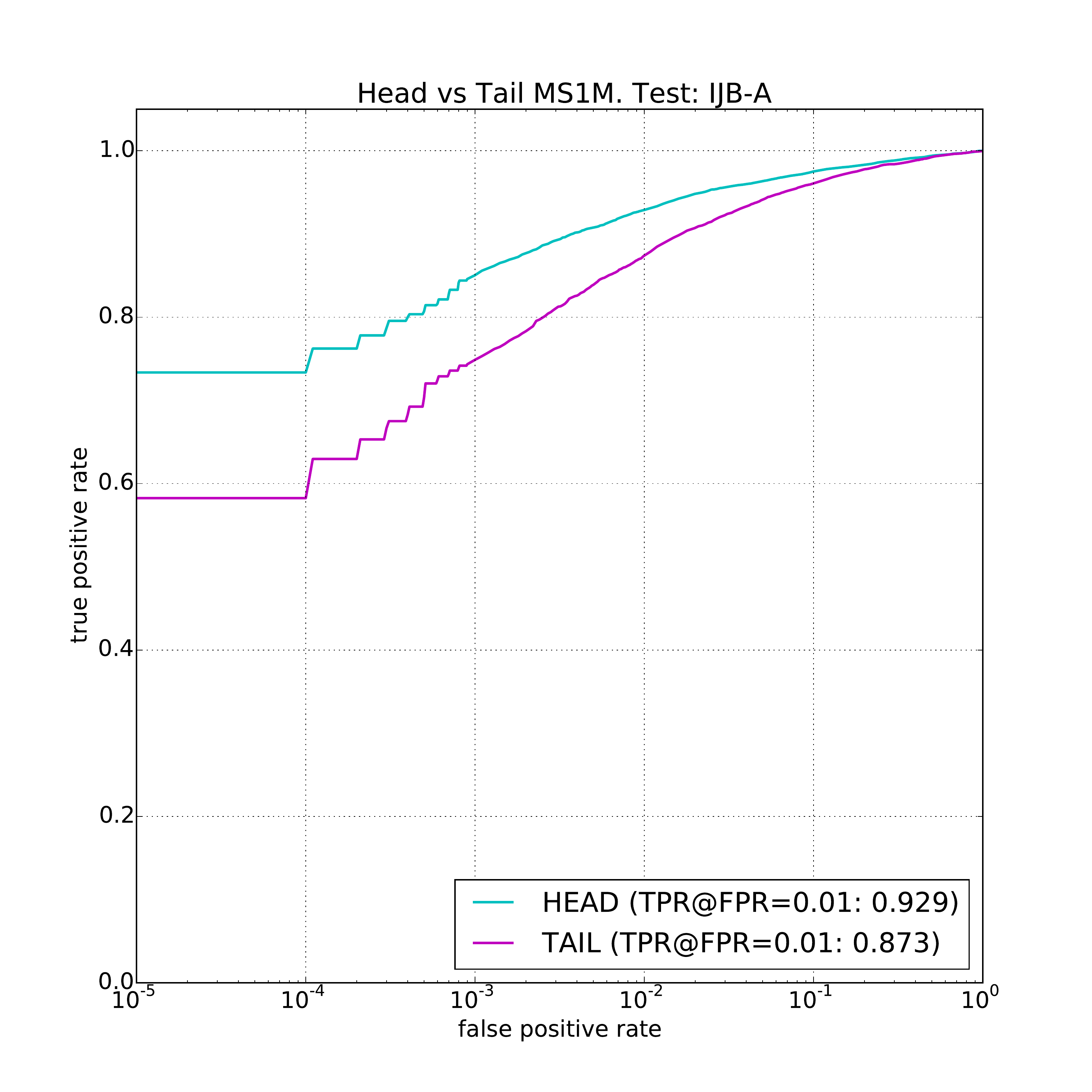}
			\caption{MS1M}
			\label{fig:myth2_resnet_ms1m}
		\end{subfigure}
		\caption{Performance on  IJB-A \cite{ijba} of ResNets trained on UMDFaces \cite{umdfaces}, and MS-Celeb-1M \cite{msceleb} `head' and `trail' sets. The `head' sets are better.}
		\label{fig:myth2_resnet}
	\end{figure}

	On the other hand, if we train deeper networks, the performance of networks trained on the `head' sets is better than the corresponding network trained on the `tail' sets. This can be seen in figure \ref{fig:myth2_resnet} where we train ResNet-101 \cite{resnet} networks on the `head' and `tail' sets of UMDFaces \cite{umdfaces} and MS-Celeb-1M \cite{msceleb} datasets and test on the IJB-A protocol \cite{ijba}.
	
	This observation is important because it can guide researchers towards better practices to follow while collecting data or selecting data for training deep networks. Data acquisition is an expensive and time consuming process and these experiments shine a light on how to obtain the maximum benefit from the investment.  This is an interesting direction for future work.

	\subsection{Does some amount of label noise help improve the performance of deep recognition networks?}
	\label{sec:myth3}
	In face identification and verification research, the effect of label noise in the training set for deep networks has not been studied extensively \cite{msceleb, msceleb1}. Label noise essentially means that some of the images have incorrect labels. Some \cite{vgg, msceleb} have suggested that deep networks are robust to label noise. 
	
	We again use CASIA-WebFace \cite{casia} and UMDFaces \cite{umdfaces} batch-1 and batch-2 for training the networks and LFW \cite{lfw}, IJB-A \cite{ijba}, UMDFaces batch-3 for evaluating the performance of these trained networks. We use the protocol explained in section \ref{sec:protocol} for evaluation. 
	
	For both training datasets, we train recognition networks with $0, 2\%, 5\%,$ and $10\%$ label noise in the dataset. We would like to point out these percentages assume that the original datasets do not already contain any label noise. This assumption might not be true for many face datasets like MS-Celeb \cite{msceleb} and VGG-Face \cite{vgg} which already contain some label noise. 
	
\begin{figure}[t]
		\begin{center}
			\includegraphics[width=\linewidth,height=0.7\linewidth]{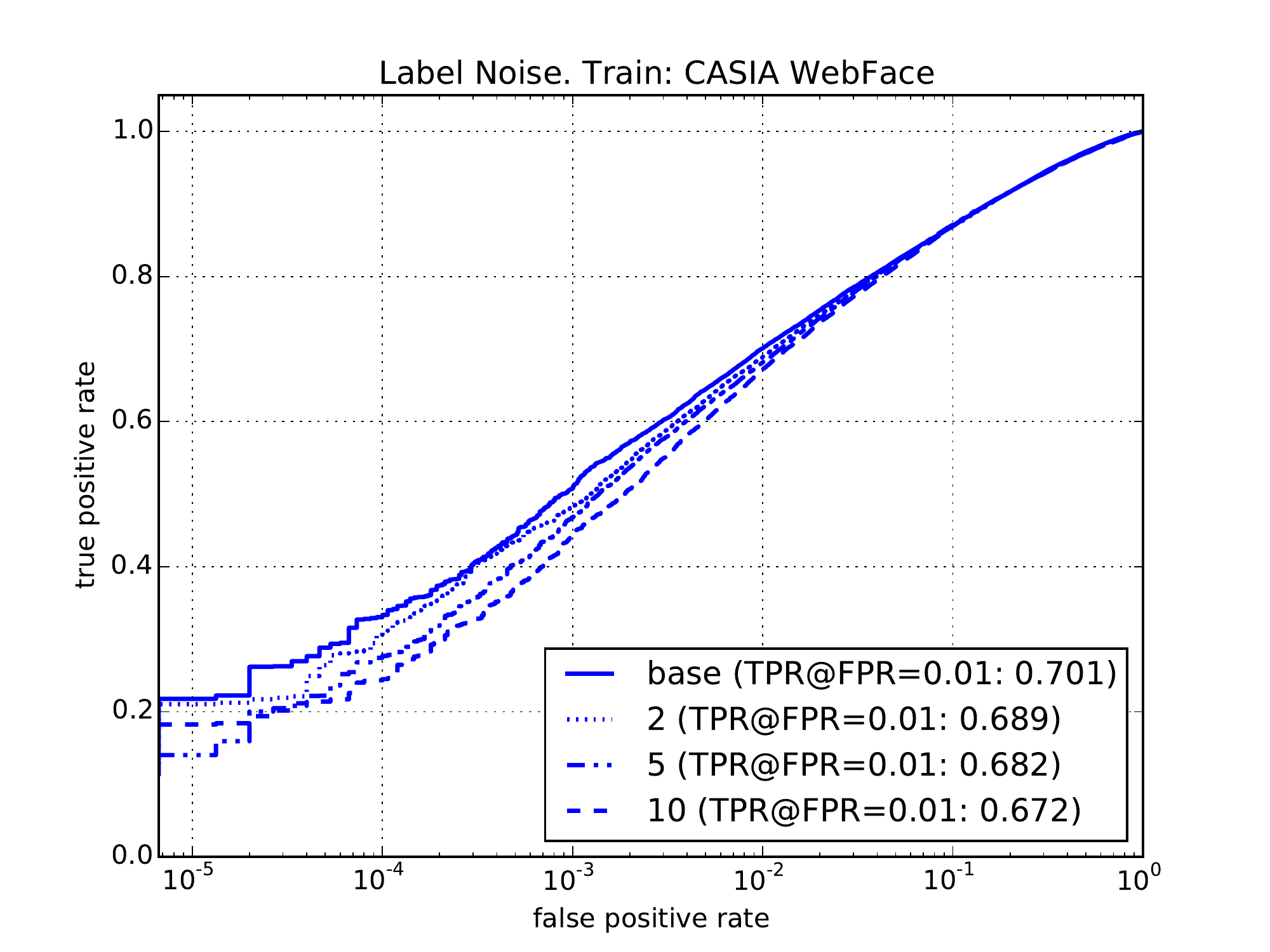}
		\end{center}
		\caption{Performance on UMDFaces batch-3 for networks trained on CASIA WebFace \cite{casia}. Similar to figure \ref{fig:myth3_umd_umd} the network trained with no label noise performs best.}
		\label{fig:myth3_casia_umd}
	\end{figure}

\begin{figure}[t]
		\begin{center}
			\includegraphics[width=\linewidth, height=0.7\linewidth]{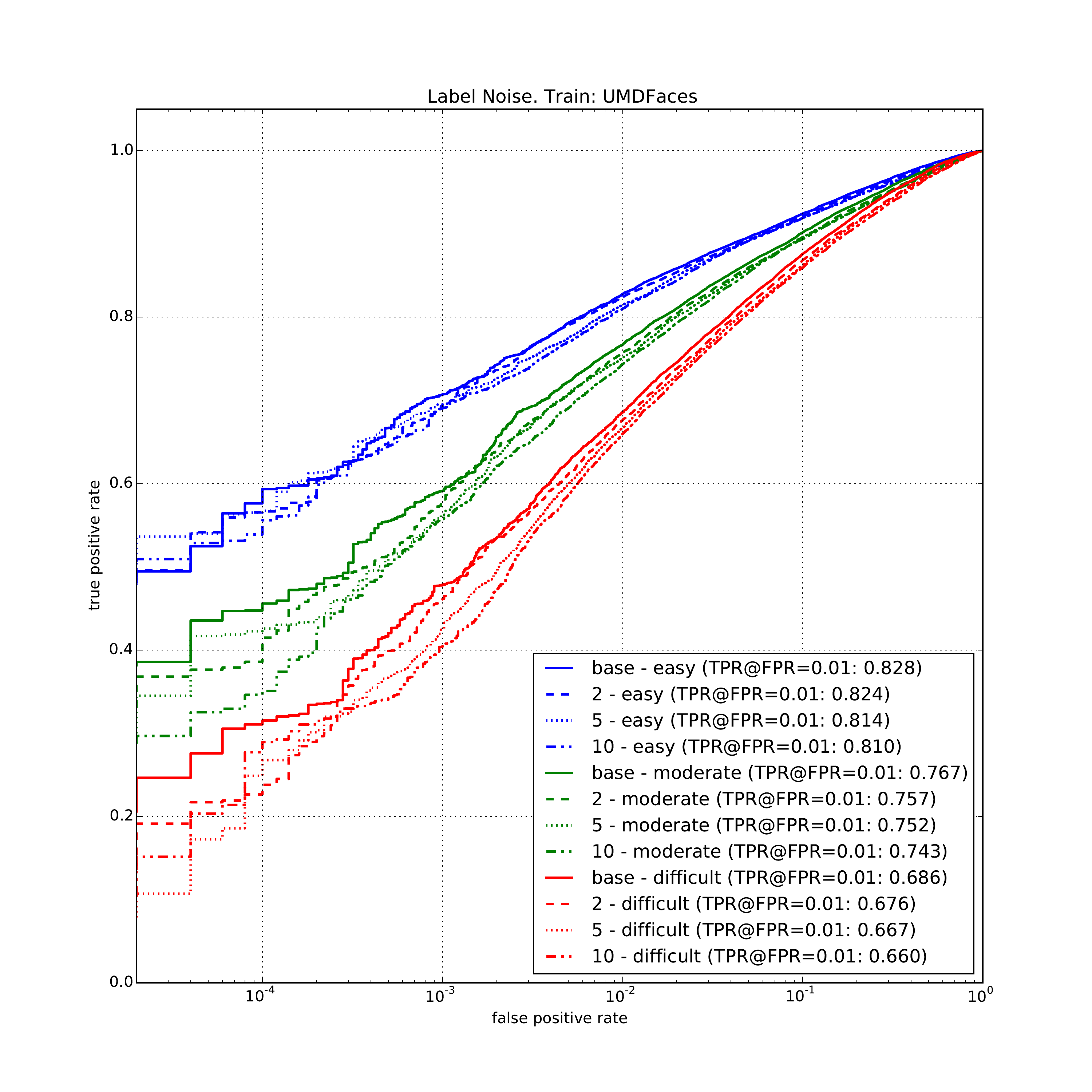}
		\end{center}
		\caption{Verification performance on UMDFaces \cite{umdfaces} batch-3 of deep networks trained on batch-1 and batch-2 with different noise levels. The colors represent the difficulty of test set (in terms of the difference in pose). Different line types represent different amounts of label noise added to the train set. Except for a small region in easy cases, using clean data is better than using data with label noise.}
		\label{fig:myth3_umd_umd}
	\end{figure}

	Figure \ref{fig:myth3_casia_umd} shows the verification performance of networks trained on the CASIA-WebFace dataset for the UMDFaces test set and figure \ref{fig:myth3_umd_umd} shows the same for networks trained on UMDFaces dataset. There is a clear degradation in performance with increasing noise level. For both datasets, the performance of the network trained on clean data is mostly better than the performance of networks trained with even small amounts of noise. Label noise does not improve performance over clean data for face recognition. However, the difference in performance between networks trained on clean data and data with low levels of label noise is relatively low. But the percentage of noisy labels should be relatively low (less than $5\%$) because from figures \ref{fig:myth3_casia_umd} and \ref{fig:myth3_umd_umd}, we notice that for a label noise level of $10\%$, the performance invariably declines. 
	
	\begin{figure}[htp]
		\centering
		\begin{subfigure}{.5\linewidth}
			\centering
			\includegraphics[width=\linewidth,height=0.8\linewidth]{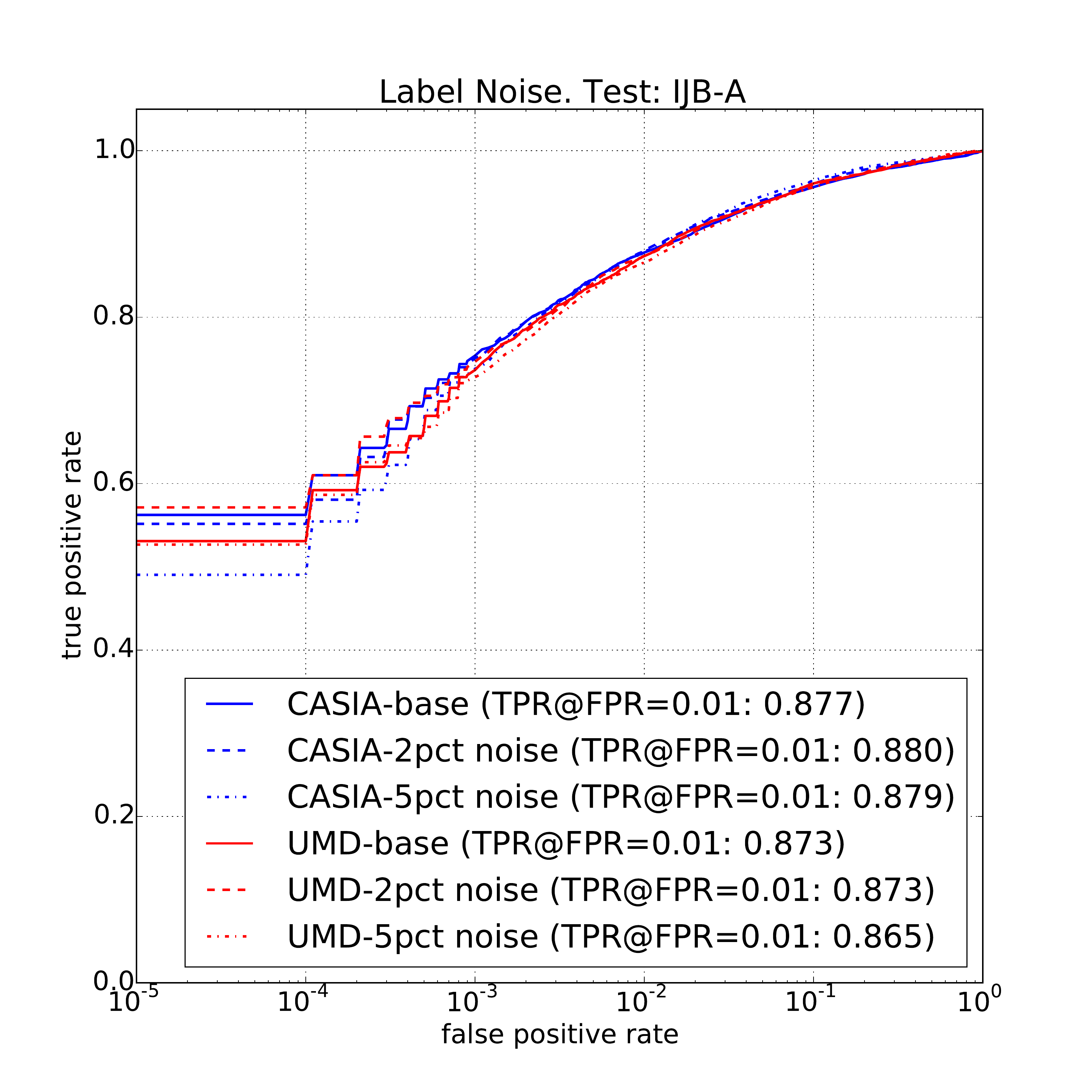}
			\caption{UMDFaces}
			\label{fig:myth3_ijba}
		\end{subfigure}%
		\begin{subfigure}{.5\linewidth}
			\centering
			\includegraphics[width=\linewidth,height=0.8\linewidth]{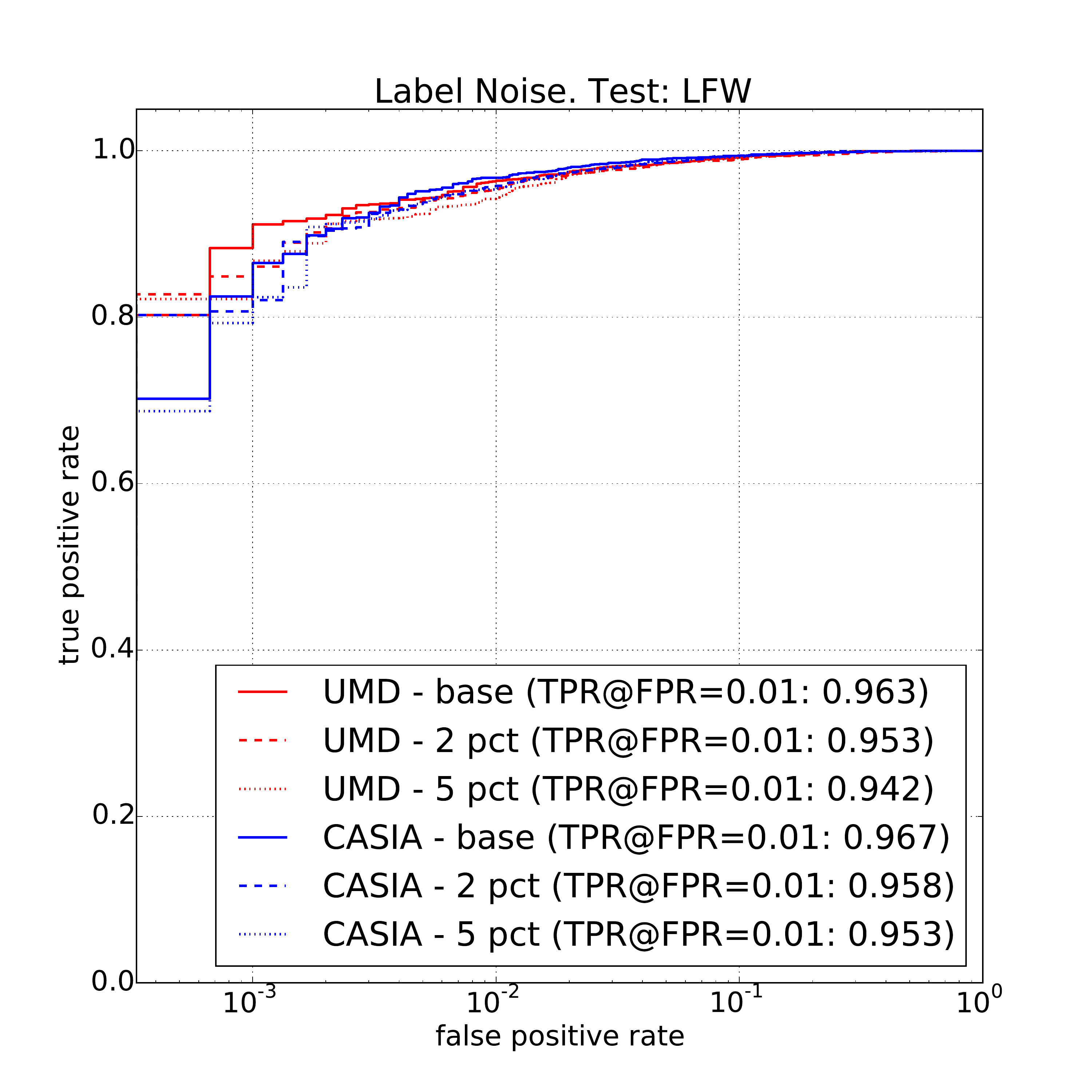}
			\caption{CASIA}
			\label{fig:myth3_lfw}
		\end{subfigure}
		\caption{Verification results on IJB-A \cite{ijba} of networks trained on (a) UMDFaces \cite{umdfaces}, and (b) CASIA WebFace \cite{casia}. Contrary to earlier observations from figures \ref{fig:myth3_umd_umd} and \ref{fig:myth3_casia_umd}, the performance on IJB-A seems to improve with adding some label noise to the train dataset. (Best viewed digitally)}
		\label{fig:myth3_ijba_lfw}
	\end{figure}
	
	Similar trends can be seen for the LFW dataset in figure \ref{fig:myth3_lfw}. However, when testing on the IJB-A dataset \cite{ijba}, we notice that this observation does not hold, as shown in figure \ref{fig:myth3_ijba}. We believe that this is because the IJB-A protocol comprises of video frames which introduces another dimension of complexity for the model. Sometimes these video frames might not look like the person under consideration. We believe that such frames might be acting like a kind of label noise in the test set. That is why adding label noise to the training set might make the networks robust to such frames. However, label noise and its removal are definitely problems worthy of further research.
	
	\subsection{Does thumbnail creation method affect performance?}
	\label{sec:myth4}
	%\cite{umdfaces,ijba,lfw,dlib,crosswhite}
	Detecting \cite{tiny, hyperface, wider, fddb, gangdet, violajones}, cropping, and aligning the faces in the dataset is the first step in many face recognition pipelines. Alignment is the process of transforming a face into some canonical view. This is usually done by detecting locations of keypoints \cite{dlib, ultraface} in the face image and then using some kind of similarity transform to transform the faces to a canonical view \cite{vgg}. We refer to the images of faces obtained after cropping and/or alignment as `thumbnails'.
	
	%The authors in \cite{vgg} suggest that aligning training images does not improve identification performance, but aligning test images does. 
	We investigate whether the performance of deep recognition networks is affected by the thumbnail generation process. We compare two popular alignment techniques \cite{ultraface, dlib} against very simple thumbnail generation techniques which only require keypoint locations and do not involve calculating any similarity transforms. 
	
\begin{figure}[t]
		\begin{center}
			\includegraphics[width=\linewidth,height=0.8\linewidth]{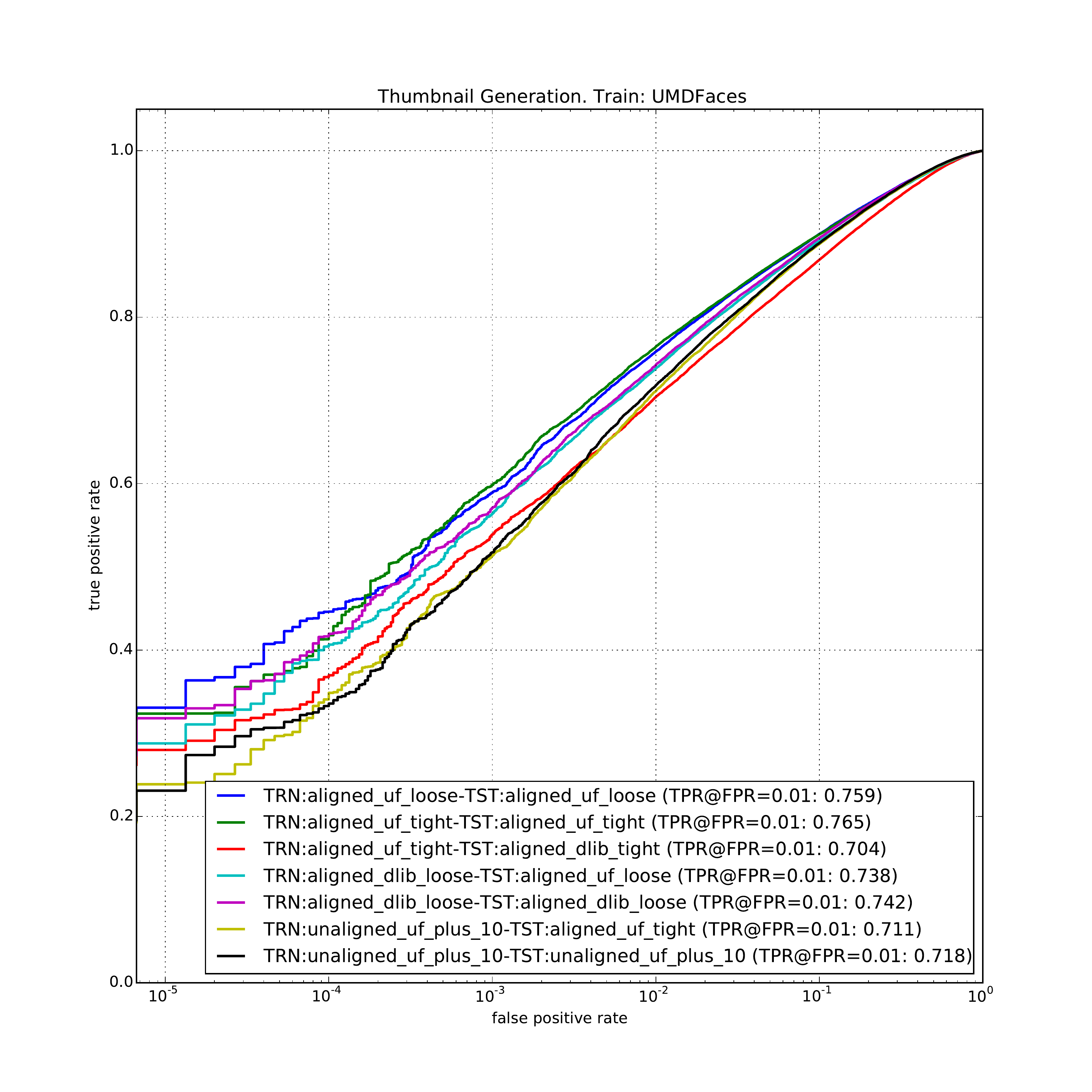}
		\end{center}
	\vspace{-7mm}
		\caption{The performance of seven sets of train and test thumbnail generation methods. These seven were selected among all pairs of train-test pairs possible as explained in section \ref{sec:myth4}. The training set was the UMDFaces \cite{umdfaces} dataset in each case and the testing set was batch-3 of UMDFaces. It is clear that tightly aligning both training and testing sets using the method from \cite{ultraface} gives the best performance (green). (Best viewed digitally)}
		\label{fig:myth4_umd}
	\end{figure}

\begin{figure}[t]
		\begin{center}
			\includegraphics[width=\linewidth,height=0.7\linewidth]{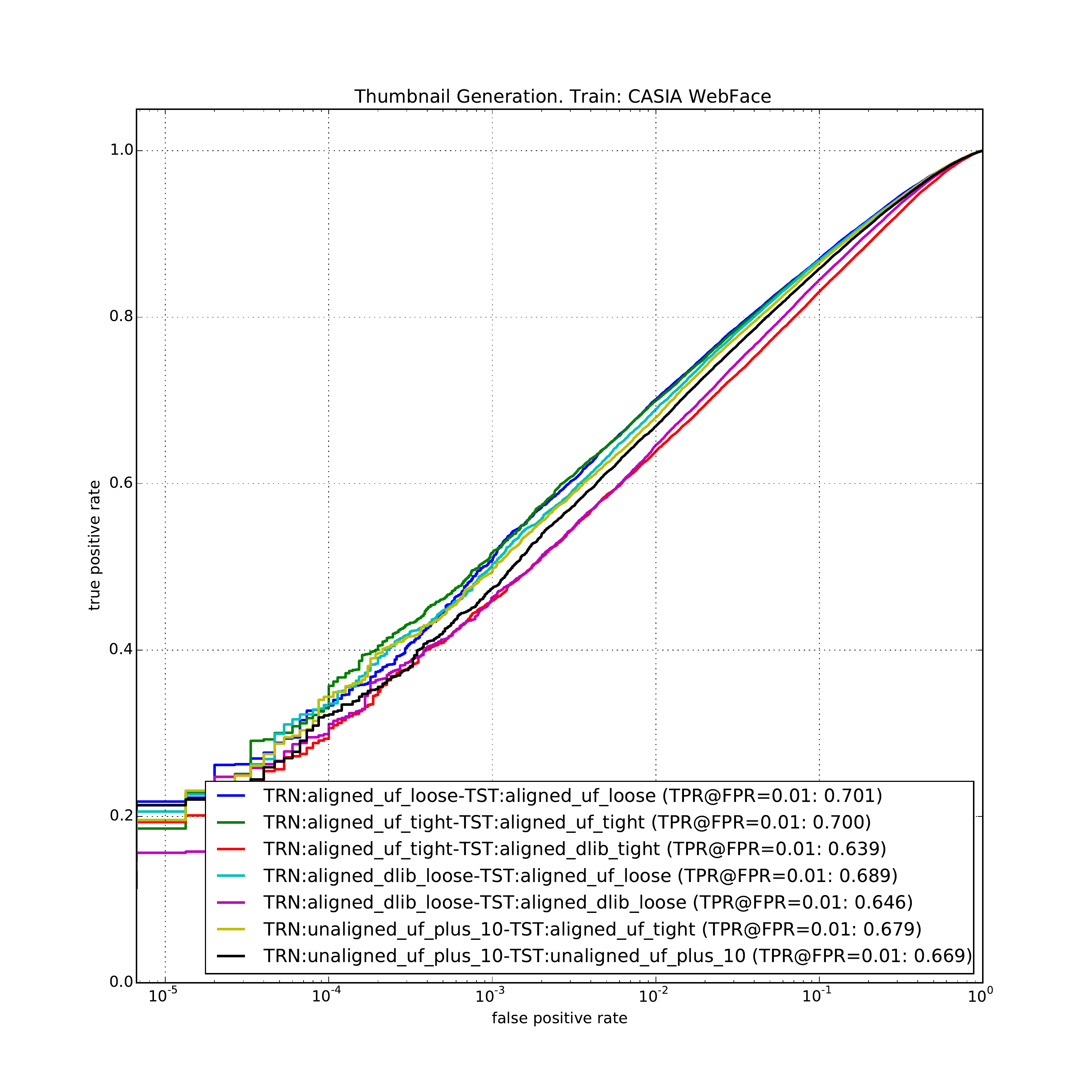}
		\end{center}
	\vspace{-7mm}
		\caption{The performance of seven sets of train and test thumbnail generation methods with CASIA WebFace as the training set and UMDFaces batch-3 \cite{umdfaces} as the test set. We again see that aligning both training and testing sets using \cite{ultraface} gives the best performance. Also, using a loose alignment gives the best performance (blue) just slightly ahead of using a tight alignment (green).}
		\label{fig:myth4_casia}
	\end{figure}

	We compare three different types of thumbnails for evaluating verification performance. These are: (i) Keypoints from All-in-one CNN \cite{ultraface} with similarity transform alignment, (ii) DLIB keypoint detection and alignment \cite{dlib}, and (iii) Bounding box using keypoints from \cite{ultraface} without any alignment. In each case, we also study the effect of using tight thumbnails (tight crop of the face) vs loose thumbnails (including more context information). We try these methods for both training and testing and present the accuracies for the best performing cases in figures \ref{fig:myth4_umd} and \ref{fig:myth4_casia}. We use two different datasets for training: batch-1 and batch-2 of UMDFaces \cite{umdfaces} and CASIA-WebFace \cite{casia}, and UMDFaces batch-3 for evaluating the performance of the trained networks. 
	
	All the above mentioned variations give us the following seven methods of thumbnail generation: (1) loose alignment using \cite{ultraface} keypoints (\textit{aligned\_uf\_loose}), (2) tight alignment using \cite{ultraface} keypoints (\textit{aligned\_uf\_tight}), (3) loose alignment using \cite{dlib} keypoints (\textit{aligned\_dlib\_loose}), (4) tight alignment using \cite{dlib} keypoints (\textit{aligned\_dlib\_tight}), (5) no alignment with extremely tight crops (max extent of the keypoints minus 10\% of the height and width from both sides) based on keypoints obtained from \cite{ultraface} (\textit{unaligned\_uf\_minus\_10}), (6) no alignment with moderately tight crops (max extent of the keypoints) based on keypoints obtained from \cite{ultraface} (\textit{unaligned\_uf\_tight}), and (7) no alignment but loose crops of the faces (max extent of the keypoints plus 10\% of the height and width on both sides) using keypoints from \cite{ultraface} (\textit{unaligned\_uf\_plus\_10}). 
	
	We train neural networks on UMDFaces \cite{umdfaces} and CASIA-WebFace \cite{casia} using these 7 thumbnail generation methods and test on batch-3 of UMDFaces \cite{umdfaces} using the same 7 different thumbnail generation methods. Hence, for both training sets, we have 49 ($7\times7$) pairs of train and test sets. For both training sets, we select the seven pairs which give the highest performance and plot them in figures \ref{fig:myth4_umd} and \ref{fig:myth4_casia}. We note that there is a clear dependence of performance on the type of thumbnail used for training and testing. Using a good keypoint detection method and alignment procedure for both training and testing is essential for achieving good performance. Note that using a tight alignment using keypoints detected using \cite{ultraface} for both training and testing gives the best performance among all the cases of networks trained on UMDFaces. This pair is also a very close second among networks trained on CASIA-WebFace. As keypoint detection and alignment methods continue to improve, we expect the face verification performance to improve too. 
	
	\section{Conclusion}
	\label{sec:conclusions}
	In this work we studied the effects of certain decisions about datasets and the training procedures for training deep convolutional neural networks for face verification. Carefully making these decisions is important for developing face recognition systems. This paper provides some guidelines about the decision making process. There is an abundance of video data which contain much more pose and expression variations than still images. To ensure that researchers can take advantage of this potential, we introduced a new dataset of \numVideo~videos and \numFrame~annotated frames. The importance of removing label noise from the dataset and selecting wider or deeper datasets cannot be ignored. Similarly, aligning faces using accurate keypoints during both training and testing gives a boost in performance. We hope that this work will encourage people to dig deeper into these and other decisions.
	
	\section*{Acknowledgments}
	This research is based upon work supported by the Office of the
	Director of National Intelligence (ODNI), Intelligence Advanced
	Research Projects Activity (IARPA), via IARPA R\&D Contract No.
	2014-14071600012. The views and conclusions contained herein are
	those of the authors and should not be interpreted as necessarily
	representing the official policies or endorsements, either expressed
	or implied, of the ODNI, IARPA, or the U.S. Government. The U.S.
	Government is authorized to reproduce and distribute reprints for
	Governmental purposes notwithstanding any copyright annotation
	thereon.
	
	{\small
		\bibliographystyle{ieee}
		\bibliography{egbib}
	}
	
%	\newpage
%	\includepdf[pages={1,2,3}]{egpaper_supplementary}
	
\end{document}